\documentclass[letterpaper, 10 pt, conference]{ieeeconf}  % Comment this line out if you need a4paper

\IEEEoverridecommandlockouts                              % This command is only needed if 
\usepackage{graphics} % for pdf, bitmapped graphics files
\usepackage{epsfig} % for postscript graphics files
\usepackage{mathptmx} % assumes new font selection scheme installed
\usepackage{times} % assumes new font selection scheme installed
\usepackage{amsmath} % assumes amsmath package installed
\usepackage{amssymb}  % assumes amsmath package installed
\usepackage{amsfonts}
\usepackage{algorithmic}
\usepackage{algorithm}
\usepackage{array}
\usepackage[caption=false,font=normalsize,labelfont=sf,textfont=sf]{subfig}
\usepackage{textcomp}
\usepackage{stfloats}
\usepackage{url}
\usepackage{verbatim}
\usepackage{graphicx}
\usepackage{cite}
\usepackage{booktabs}
\usepackage{threeparttable}
\usepackage{makecell}
\usepackage{color}
\usepackage{xcolor}
\usepackage{multicol}
\usepackage{multirow}
\usepackage{soul}
\usepackage{hyperref}

\title{\LARGE \bf
TopoDiffuser: A Diffusion-Based Multimodal Trajectory Prediction Model with Topometric Maps
}

\author{
Zehui Xu\textsuperscript{1*},
Junhui Wang\textsuperscript{2,3*},
Yongliang Shi\textsuperscript{2},
Chao Gao\textsuperscript{2$\dagger$},
Guyue Zhou\textsuperscript{2,4$\dagger$}
\thanks{* Equal contribution.}
\thanks{$\dagger$ Corresponding author: Chao Gao and Guyue Zhou.}
\thanks{\textsuperscript{1}School of Astronautics, Harbin Institute of Technology, \textsuperscript{2}Institute for AI Industry Research (AIR), Tsinghua University, 
\textsuperscript{3}Institute of Systems Engineering and Collaborative Laboratory for Intelligent Science and Systems, Macau University of Science and Technology, $^{4}$School of Vehicle and Mobility, Tsinghua University.}
\thanks{Sponsored by Xinchen Qihang Inc.}
}

\begin{document}

\maketitle
\thispagestyle{empty}
\pagestyle{empty}

%%%%%%%%%%%%%%%%%%%%%%%%%%%%%%%%%%%%%%%%%%%%%%%%%%%%%%%%%%%%%%%%%%%%%%%%%%%%%%%%
\begin{abstract}
This paper introduces TopoDiffuser, a diffusion-based framework for multimodal trajectory prediction that incorporates topometric maps to generate accurate, diverse, and road-compliant future motion forecasts. By embedding structural cues from topometric maps into the denoising process of a conditional diffusion model, the proposed approach enables trajectory generation that naturally adheres to road geometry without relying on explicit constraints. A multimodal conditioning encoder fuses LiDAR observations, historical motion, and route information into a unified bird’s-eye-view (BEV) representation. Extensive experiments on the KITTI benchmark demonstrate that TopoDiffuser outperforms state-of-the-art methods, while maintaining strong geometric consistency. Ablation studies further validate the contribution of each input modality, as well as the impact of denoising steps and the number of trajectory samples. To support future research, we publicly release our code at \href{https://github.com/EI-Nav/TopoDiffuser}{https://github.com/EI-Nav/TopoDiffuser}.
\end{abstract}

%%%%%%%%%%%%%%%%%%%%%%%%%%%%%%%%%%%%%%%%%%%%%%%%%%%%%%%%%%%%%%%%%%%%%%%%%%%%%%%%
\section{Introduction}
Trajectory prediction is an important task in autonomous driving and robotic navigation. It helps intelligent agents anticipate the future movements of surrounding vehicles, pedestrians, and other dynamic objects, enabling safer planning. However, driving behavior is uncertain and varies in different situations. For the same past motion and environment, there can be multiple possible future paths. This makes trajectory prediction a challenging problem.

To be effective in complex and dynamic traffic environments, a trajectory prediction model should not only generate accurate motion forecasts that comply with road geometry, but also capture the inherent multi-modality of future behaviors. The ability to produce diverse trajectory hypotheses is essential for downstream planning and decision-making, particularly in the presence of uncertainty and interaction among agents.

Topometric maps provide rich semantic and geometric cues that are instrumental in guiding the prediction of feasible and road-compliant trajectories. However, existing prediction models often fail to fully exploit this structured information while preserving the flexibility required to represent the multi-modal nature of future agent behaviors.

To address this challenge, we propose TopoDiffuser, a novel diffusion-based multimodal trajectory prediction model that leverages topometric maps as guidance in the diffusion process. Recent advances in diffusion models have demonstrated their strong capability in capturing complex multimodal distributions, making them well-suited for trajectory forecasting. However, existing diffusion-based approaches often lack explicit spatial constraints, leading to predicted trajectories that may deviate from feasible driving paths. In TopoDiffuser, we incorporate topometric maps as a guiding mechanism within the diffusion process, ensuring that generated trajectories remain both diverse and road-compliant without explicitly enforcing hard constraints.

We validate TopoDiffuser through extensive experiments on large-scale trajectory prediction benchmarks, demonstrating its superiority over state-of-the-art methods. Our key contributions are as follows.

\begin{enumerate}
\item A diffusion-based trajectory prediction method that effectively captures the multimodal nature of future driving behaviors.

\item The use of off-the-shelf topometric maps as guidance in the diffusion process, ensuring that predicted trajectories remain feasible and aligned with road structures while preserving diversity.

\item Comprehensive evaluations on large-scale datasets, showcasing the effectiveness of TopoDiffuser in generating diverse and accurate trajectory forecasts. To facilitate further research, we open-source our code at \href{https://github.com/EI-Nav/TopoDiffuser}{https://github.com/EI-Nav/TopoDiffuser}.
\end{enumerate}

\begin{figure}[!t]
\centering
\includegraphics[width=0.48\textwidth]{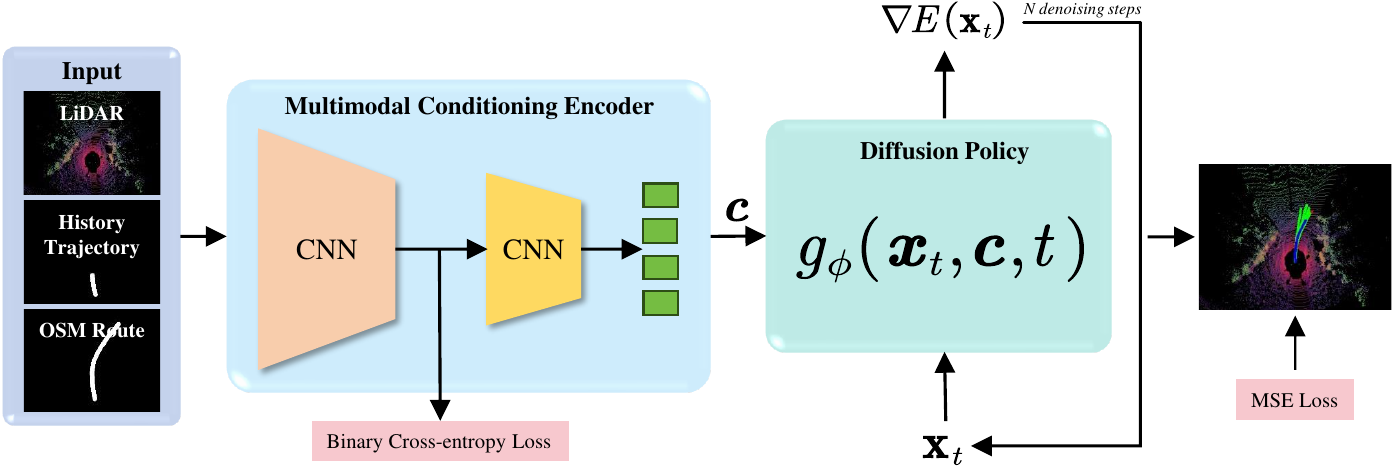}%
\caption{Overview of the proposed diffusion-based trajectory prediction framework.}
\label{fig:main}
\vspace{-4mm}
\end{figure}

\section{Related Works}
\subsection{Deep Learning-Based Trajectory Prediction}
Deep learning has advanced trajectory prediction by capturing complex spatiotemporal dependencies in dynamic environments. Early approaches primarily relied on Recurrent Neural Networks (RNNs), particularly Long Short-Term Memory (LSTM) models \cite{Chang2019,Deo2018,Ding2019}, to model sequential motion patterns. However, RNN-based models often struggle with long-term dependencies and suffer from vanishing gradients.

To address these limitations, more recent works have adopted Temporal Convolutional Networks (TCNs) \cite{Katariya2022,Li2022,Zhang2021} and Transformer-based architectures \cite{Chen2021,Liu2021,Quintanar2021}, which offer superior parallelism and improved capacity to model long-range temporal structures. Furthermore, interaction-aware models have gained traction, leveraging Graph Neural Networks (GNNs) \cite{Chandra2020,Li2019} and attention mechanisms \cite{Fu2022,Meng2021} to capture the influence of surrounding agents in multi-agent driving scenarios.

% Deep learning has significantly advanced trajectory prediction by effectively capturing complex spatiotemporal dependencies. Among various approaches, Recurrent Neural Networks (RNNs), particularly Long Short-Term Memory (LSTM) networks \cite{Chang2019,Deo2018,Ding2019}, have been widely used to model sequential dependencies in vehicle motion. However, RNN-based models often suffer from vanishing gradient issues and struggle to capture long-range dependencies effectively. To address these limitations, Temporal Convolutional Networks (TCNs) \cite{Katariya2022,Li2022,Zhang2021} and Transformer-based architectures \cite{Chen2021,Liu2021,Quintanar2021} have been proposed. These models offer improved performance by enabling efficient long-term dependency modeling and better handling of multimodal trajectory distributions.

% Another critical aspect of trajectory prediction is interaction modeling, as the motion of an agent is influenced by surrounding vehicles. To represent these interactions, Graph Neural Networks (GNNs) \cite{Chandra2020,Gindele2015,Li2019} have been employed, where vehicles are treated as graph nodes and their interactions are captured through edge connections. Additionally, attention mechanisms \cite{Fu2022,Kim2017,Meng2021} have been incorporated to dynamically weigh the influence of surrounding agents, improving prediction accuracy in complex multi-agent driving scenarios.

\subsection{Generative Models for Multimodal Prediction}
Trajectory prediction is inherently multimodal, as multiple future behaviors may be plausible under the same historical context. Deterministic models often fail to capture this uncertainty, prompting the use of generative approaches. Generative Adversarial Networks (GANs) \cite{Guo2023,Hegde2020,Li2021} enable the generation of diverse trajectories but are prone to mode collapse. Variational Autoencoders (VAEs) \cite{Bhattacharyya2020,Cho2019,Dulian2021} provide better coverage of the trajectory distribution but often yield overly smoothed predictions lacking sharpness in maneuvers.

More recently, denoising diffusion probabilistic models have emerged as a powerful alternative for multimodal prediction \cite{Bahram2016}. These models progressively refine noisy samples into coherent trajectories and have demonstrated strong performance in representing distributional uncertainties.

% Trajectory prediction is inherently multimodal, as agents may follow multiple plausible future paths under the same initial conditions. Traditional deterministic models often struggle to represent this uncertainty effectively. To address this challenge, Generative Adversarial Networks (GANs) \cite{Guo2023,Hegde2020,Li2021} have been explored, where a generator produces realistic trajectory samples while a discriminator differentiates them from ground truth. However, GAN-based methods often suffer from mode collapse, leading to reduced trajectory diversity.

% Variational Autoencoders (VAEs) \cite{Bhattacharyya2020,Cho2019,Dulian2021} provide an alternative by learning a latent space representation of trajectory distributions. While VAEs can generate diverse trajectory samples, they often produce overly smoothed predictions that fail to capture sharp maneuvers. More recently, Diffusion Models have gained attention for trajectory prediction due to their ability to generate diverse and realistic future trajectories by progressively refining noisy samples into coherent paths \cite{Bahram2016}. These models have demonstrated superior performance in capturing multimodal uncertainties and improving prediction robustness compared to GANs and VAEs.

\subsection{Topometric Map-Based Approaches}
High-definition and topometric maps provide rich semantic and geometric context essential for road-compliant motion forecasting. Prior works have explored map integration via graph-based representations \cite{Gao2020,Gu2021,Liang2020,Zeng2021} and lane-aware attention mechanisms \cite{Wang2023}, enabling predictions that better conform to road topology. Vectorized and rasterized map encodings have also been used to support spatial reasoning in learning-based models.

% Nonetheless, many existing methods struggle to jointly preserve multimodal diversity and geometric feasibility. Our work addresses this gap by embedding topometric map information directly into the diffusion process, enabling diverse yet road-aligned trajectory generation without relying on hard constraints.

% Road topology plays a fundamental role in constraining and guiding vehicle motion. To enhance trajectory prediction accuracy, recent works have integrated high-definition (HD) maps and topometric representations \cite{Gao2020,Gu2021,Liang2020}. Graph-based road representations \cite{Zeng2021} and lane-aware attention mechanisms \cite{Wang2023} have been employed to align predicted trajectories with feasible road structures. Additionally, vectorized map representations \cite{Gu2021} enable efficient spatial reasoning and allow learning-based models to incorporate road constraints more effectively.

Despite these advancements, existing methods often struggle to jointly capture both multimodal trajectory distributions and the influence of road topology. To address this gap, TopoDiffuser integrates a diffusion-based generative framework with topometric map embeddings, ensuring that predicted trajectories are both diverse and physically plausible within real-world driving constraints.

\section{Proposed Method}
This work addresses the problem of multimodal trajectory prediction for robots. Given LiDAR observations, the robot's ego-motion history, and guidance from a topometric map, the objective is to estimate a distribution over future trajectories that are both diverse and physically plausible.
\subsection{Problem Formulation}
Let $\tau = \{x_1, \dots, x_{T_f}\} \in \mathbb{R}^{T_f \times 2}$ represent a future trajectory over $T_f$ time steps. Let $c$ denote the spatiotemporal context derived from the fused bird’s-eye view (BEV) representation of LiDAR data, past motion, and road topology. The objective is to model the conditional distribution $p_\phi(\tau \mid c)$ that captures the uncertainty inherent in future motion.

To approximate this distribution, we employ a conditional denoising diffusion process. The forward process progressively perturbs the ground-truth trajectory $\tau_0$ by adding Gaussian noise.
\begin{equation}
    q(\tau_t \mid \tau_0) = \mathcal{N}(\tau_t; \sqrt{\gamma_t} \tau_0, (1 - \gamma_t)\mathbf{I}),
\end{equation}
where $\gamma_t$ is a monotonically decreasing noise schedule. The reverse process is parameterized by a neural network $g_\phi$, which reconstructs $\tau_0$ from the noisy sample $\tau_t$, conditioned on the context $c$ and diffusion step $t$.

The overall architecture of the proposed network is illustrated in Fig.~\ref{fig:main}. The model is trained to minimize the reconstruction error between the predicted and true noise, enabling the generation of diverse trajectory samples that conform to both the semantic scene structure and the physical constraints of the driving environment.

\subsection{Input Representation}
The input to our model consists of three spatially aligned modalities: LiDAR point clouds, ego-trajectory history, and topometric map guidance. All inputs are rasterized into a unified BEV frame with resolution $H_0 \times W_0$, centered on the ego vehicle.

Formally, the three components are encoded as follows.

\begin{itemize}
    \item \textit{LiDAR BEV Encoding:} Following~\cite{Zeng2019}, raw LiDAR point clouds are projected onto a BEV grid and encoded as a tensor $\mathbf{I}_{\text{lidar}} \in \mathbb{R}^{H_0 \times W_0 \times 3}$, where the three channels correspond to height, intensity, and point density.
    
    \item \textit{Trajectory History Encoding:} The past trajectory of the ego vehicle over $T_h$ frames is rasterized into a binary occupancy map $\mathbf{I}_{\text{traj}} \in \mathbb{R}^{H_0 \times W_0 \times 1}$, where each occupied pixel indicates a historical position.

    \item \textit{Topometric Map Encoding:} The sparse topometric route derived from OpenStreetMap (OSM) is converted into a binary mask $\mathbf{I}_{\text{map}} \in \mathbb{R}^{H_0 \times W_0 \times 1}$ that indicates the feasible driving corridor in the local BEV space.
\end{itemize}

These three components are concatenated along the channel dimension to form the final input tensor:
\begin{equation}
    \mathbf{I}_{\text{input}} = \text{Concat}(\mathbf{I}_{\text{lidar}}, \mathbf{I}_{\text{traj}}, \mathbf{I}_{\text{map}}) \in \mathbb{R}^{H_0 \times W_0 \times 5}
\end{equation}

\subsection{Multimodal Conditioning Encoder}
To integrate the heterogeneous input modalities, we propose a multimodal conditioning encoder. An overview of the encoder architecture is provided in Fig.~\ref{fig:main}. It transforms the composite input tensor $\mathbf{I}_{\text{input}}$ into a context-aware representation that serves as the conditioning signal for trajectory generation.

The encoder consists of two stages. In the first stage, inspired by the method proposed in~\cite{Xu2022}, we employ a convolutional neural network (CNN) backbone to extract hierarchical BEV features while simultaneously predicting a road segmentation mask. This backbone produces an intermediate feature map $\mathbf{F}_{\text{CNN}} \in \mathbb{R}^{H_1 \times W_1 \times 1}$, which captures both semantic and geometric information from the environment.

In the second stage, the predicted segmentation mask is further processed by an additional CNN block, yielding a compact feature map $\mathbf{F}_{\text{cond}} \in \mathbb{R}^{H_2 \times W_2 \times 1}$. This tensor is subsequently reshaped into a vector $\mathbf{c} \in \mathbb{R}^{H_2 W_2}$, which serves as the final conditioning signal.

This architectural design enables the encoder to effectively fuse low-level spatial cues with high-level semantic understanding of the roadway layout. The road segmentation module is trained using a binary cross-entropy loss computed between the predicted mask and the ground-truth virtual road mask. The resulting vector $\mathbf{c}$ provides a compact multimodal context representation, which is subsequently used as the observation condition for diffusion-based trajectory prediction.

\subsection{Conditional Diffusion Model}
We adopt a conditional denoising diffusion framework to model the distribution $p_\phi(\tau \mid c)$ over future trajectories $\tau \in \mathbb{R}^{T_f \times 2}$. Following the Diffusion Policy paradigm, the model iteratively refines a noisy trajectory $\tau_t$ into a feasible prediction $\tau_0$, conditioned on the scene context $c$.

The denoising function $g_\phi$, implemented as a lightweight CNN, predicts the noise component $\hat{\varepsilon}_t$ at each diffusion step $t$, given the inputs $\tau_t$, $t$, and $c$. The timestep is embedded using sinusoidal positional encodings, while the context $c$ is derived from the multimodal conditioning encoder, which fuses BEV features from LiDAR, past trajectories, and topometric maps.

This conditional generation process enables the model to produce denoised trajectories that are diverse, realistic, and aligned with road geometry, without requiring explicit constraint enforcement.

\subsection{Training and Inference}
The proposed \textit{TopoDiffuser} learns to generate future trajectories via a conditional diffusion model. At each denoising step, the model predicts the noise component added to the clean trajectory sample. The diffusion process is supervised using a mean squared error (MSE) loss between the predicted and actual noise, defined as

\begin{equation}
\mathcal{L}_{\text{diffusion}} = \mathbb{E}_{t, \tau_0, \varepsilon} \left[ \left\| \varepsilon - g_\phi(\tau_t, t, c) \right\|^2 \right]
\label{eq:diffusion_loss}
\end{equation}
where $\tau_0$ is the ground-truth trajectory, $\tau_t$ is the noisy trajectory at timestep $t$, $\varepsilon$ is the sampled Gaussian noise, $c$ is the conditioning context, and $g_\phi$ is the denoising network parameterized by $\phi$.

To enhance road-awareness, we use an auxiliary road segmentation head trained to predict a probability map of drivable areas. The ground-truth segmentation mask is constructed by rasterizing the recorded driving trajectory into a binary image $\mathbf{y} \in \{0,1\}^{H' \times W'}$, where each pixel indicates whether it belongs to the traversed road region. The predicted road mask is denoted as $\mathbf{x} \in [0,1]^{H' \times W'}$. The segmentation loss is defined as a pixel-wise binary cross-entropy.

\begin{equation}
\mathcal{L}_{\text{road}} = -\sum_{i=1}^{H'} \sum_{j=1}^{W'} \left[ \mathbf{y}_{i,j} \log(\mathbf{x}_{i,j}) + (1 - \mathbf{y}_{i,j}) \log(1 - \mathbf{x}_{i,j}) \right]
\label{eq:road_loss}
\end{equation}

The final training objective combines both losses.

\begin{equation}
\mathcal{L}_{\text{total}} = \mathcal{L}_{\text{diffusion}} + \lambda_{\text{road}} \cdot \mathcal{L}_{\text{road}},
\end{equation}
where $\lambda_{\text{road}}$ is a weighting factor that balances the contribution of road segmentation supervision.

During inference, only the trained diffusion model is used to generate trajectory samples. The road segmentation head is discarded, ensuring that the auxiliary supervision does not affect runtime performance.

% During training, we follow the standard denoising diffusion objective. Given a ground-truth trajectory $\tau_0$, we sample a diffusion timestep $t \sim \mathcal{U}(1, T)$ and corrupt $\tau_0$ with Gaussian noise to obtain a noisy sample $\tau_t$. The denoising model $g_\phi$ is trained to predict the added noise $\varepsilon_t$, minimizing the mean squared error between the predicted and true noise.
% \begin{equation}
% \mathcal{L}_{\text{diff}} = \mathbb{E}_{t, \tau_0, \varepsilon} \left[ \left\| \varepsilon - g_\phi(\tau_t, t, c) \right\|^2 \right]
% \label{eq:diffusion_loss}
% \end{equation}
% where the context $c$ is computed from the multimodal conditioning encoder, and $\varepsilon \sim \mathcal{N}(0, I)$ is sampled independently at each step.

% At inference time, we start from a random Gaussian trajectory $\tau_T \sim \mathcal{N}(0, I)$ and iteratively apply the reverse diffusion process for $T$ steps using the learned model $g_\phi$. This yields a sample $\tau_0$ that is coherent with the scene and compliant with road and motion constraints, as encoded by the context $c$.

\section{Experiments} \label{sec:5-B}
\subsection{Dataset}
We conduct our experiments on the KITTI raw dataset~\cite{Geiger:2012}, which features diverse urban driving scenarios including intersections and multilane roads. Following the KITTI odometry benchmark protocol, sequences \texttt{00}, \texttt{02}, \texttt{05}, and \texttt{07} are used for training, while sequences \texttt{08}, \texttt{09}, and \texttt{10} serve as the test set. This split yields 3,860 training samples and 1,391, 530, and 349 test samples for the respective sequences.

\subsection{Implementation Details}
The model is trained on a single NVIDIA RTX 4090D GPU using the Adam optimizer with an initial learning rate of $3 \times 10^{-3}$ and cosine decay. Training is conducted for 120 epochs with a batch size of 8. The diffusion process uses 10 denoising steps, and the denoising network is implemented as a lightweight U-Net.

The input trajectory history consists of the past 5 keyframes, each spaced at 2-meter intervals. For topological guidance, we extract a route from OSM, centered at the current ego position and covering both the past and future driving paths. Specifically, the OSM route includes 5 keyframes into the past and 15 into the future, sampled every 2 meters.

During inference, we sample 5 trajectories by running the reverse diffusion process multiple times with independent Gaussian noise, allowing the model to generate diverse and plausible future motions.

\subsection{Evaluation Metrics}
To comprehensively evaluate the performance of our trajectory prediction model, we adopt four widely used metrics: Final Displacement Error (FDE), Minimum Average Displacement Error (minADE), HitRate, and Hausdorff Distance (HD). Each metric captures complementary aspects of prediction quality.

\subsubsection{Final Displacement Error}
FDE measures the Euclidean distance between the predicted final position and the ground-truth endpoint.
\begin{equation}
    \text{FDE} = \left\| \hat{s}_{T-1} - s^*_{T-1} \right\|_2
\end{equation}

This metric evaluates the model’s ability to accurately forecast the robot’s final destination, which is crucial for downstream planning tasks.

\subsubsection{Minimum Average Displacement Error}
Given a set of $k$ predicted trajectory hypotheses, minADE computes the minimum average displacement error across all candidates.
\begin{equation}
    \text{minADE}_k = \min_{\hat{s}^k} \frac{1}{T} \sum_{t=0}^{T-1} \left\| \hat{s}^k_t - s^*_t \right\|
\end{equation}

This metric reflects the model’s capacity to represent multimodal trajectories by selecting the most accurate hypothesis.

\subsubsection{HitRate}
HitRate quantifies the proportion of predicted trajectories that remain within a fixed distance threshold $d$ from the ground truth.
\begin{equation}
    \text{HitRate}_{k,d} = \frac{1}{N} \sum_{i=1}^{N} \mathbb{I} \left[ \min_{\hat{s}^k} \max_t \left\| \hat{s}^k_t - s^*_t \right\| < d \right]
\end{equation}

This metric evaluates whether at least one prediction is sufficiently close to the ground truth.

\subsubsection{Hausdorff Distance}
HD captures the worst-case geometric deviation between predicted and reference trajectories.
\begin{equation}
    \text{HD}(s, s^*) = \max \left\{ \sup_{a \in s} \inf_{b \in s^*} \left\| a - b \right\|, \sup_{b \in s^*} \inf_{a \in s} \left\| b - a \right\| \right\}
\end{equation}

This metric provides a stringent measure of spatial alignment, especially critical in safety-sensitive scenarios.

\begin{figure*}[ht]
    \centering
    \includegraphics[width=\textwidth]{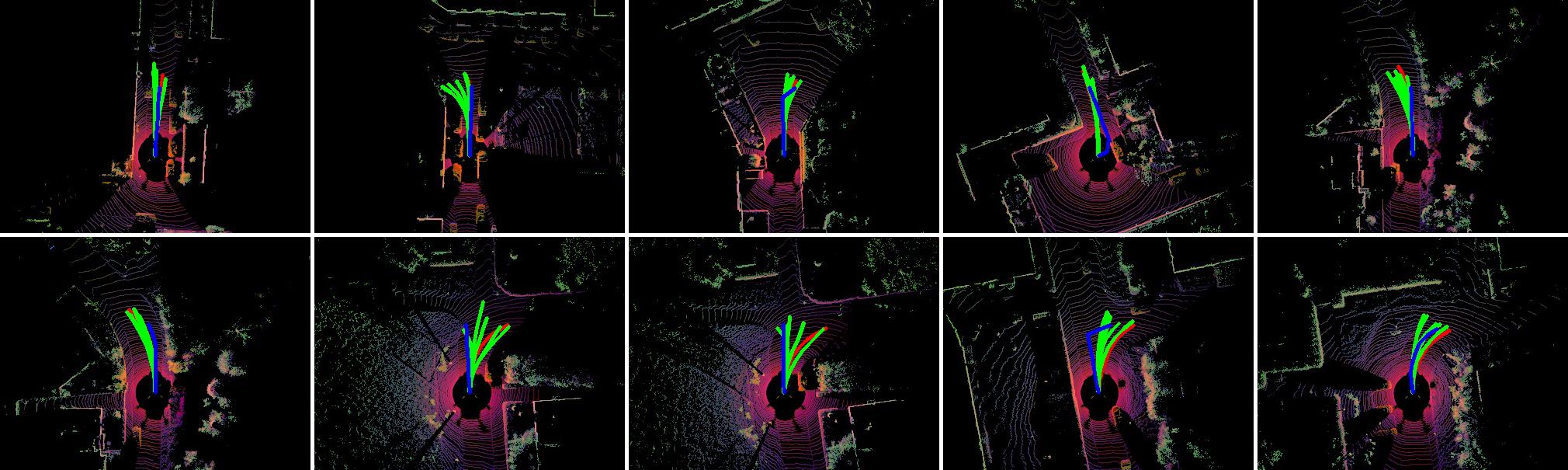}
    \caption{Predicted trajectories on representative scenes. Blue for OSM route, red for ground truth, green for predictions.}
    \label{fig:results}
\end{figure*}

\begin{table*}[ht]
    \caption{Quantitative results on the KITTI dataset.}
    \small
    \centering
    \begin{tabular}{c l c c c c c}
        \toprule
        Dataset & Method & FDE $\downarrow$ & minADE $\downarrow$ & HitRate $\uparrow$ & HD $\downarrow$ & Infer Time (s) \\
        \midrule

        \multirow{4}{*}{KITTI-08}
            & CoverNet \cite{2019CoverNet} & 4.59 & 0.59 & 0.82 & 2.58 & \textbf{0.005} \\
            & MTP \cite{2018Multimodal}    & 1.38 & 0.39 & 0.89 & 1.71 & 0.013 \\
            & TP \cite{Xu2022}             & 0.98 & 0.46 & 0.87 & 2.39 & 0.014 \\
            & \textbf{TopoDiffuser}                & \textbf{0.56} & \textbf{0.26} & \textbf{0.93} & \textbf{1.33} & 0.053 \\
        \midrule

        \multirow{4}{*}{KITTI-09}
            & CoverNet \cite{2019CoverNet} & 4.48 & 0.43 & 0.85 & 2.74 & \textbf{0.006} \\
            & MTP \cite{2018Multimodal}    & 1.07 & 0.18 & 0.98 & 1.62 & 0.013 \\
            & TP \cite{Xu2022}             & 0.55 & 0.23 & 0.98 & 2.49 & 0.018 \\
            & \textbf{TopoDiffuser}                & \textbf{0.31} & \textbf{0.13} & \textbf{0.99} & \textbf{1.21} & 0.055 \\
        \midrule

        \multirow{4}{*}{KITTI-10}
            & CoverNet \cite{2019CoverNet} & 4.33 & 0.51 & 0.85 & 2.74 & \textbf{0.007} \\
            & MTP \cite{2018Multimodal}    & 1.03 & 0.25 & \textbf{0.96} & 2.26 & 0.014 \\
            & TP \cite{Xu2022}             & 0.62 & 0.26 & 0.95 & 2.97 & 0.016 \\
            & \textbf{TopoDiffuser}                & \textbf{0.46} & \textbf{0.19} & \textbf{0.96} & \textbf{2.18} & 0.054 \\
        \bottomrule
    \end{tabular}
    \label{tab:results}
\end{table*}

\subsection{Empirical Results}

We evaluate the proposed \textit{TopoDiffuser} method on the KITTI dataset and compare its performance against several state-of-the-art baselines, including CoverNet~\cite{2019CoverNet}, MTP~\cite{2018Multimodal}, and TP~\cite{Xu2022}. 

\subsubsection{Quantitative Comparison}

Table~\ref{tab:results} summarizes the quantitative results on three KITTI sequences. Our method consistently outperforms existing approaches across all evaluation metrics.

In KITTI-08, TopoDiffuser achieves an FDE of 0.56\,m and a minADE of 0.26\,m, outperforming MTP by 33\% in FDE and 33\% in minADE. The HitRate reaches 0.93, a 4.5\% improvement over MTP, while the HD decreases to 1.33, indicating improved geometric consistency.

Performance gains are even more significant on KITTI-09, where TopoDiffuser achieves a minADE of 0.13\,m and an FDE of 0.31\,m, representing 28\% and 44\% improvements over the next best method, respectively. The HitRate reaches 0.99 and HD is reduced to 1.21, demonstrating both predictive accuracy and road compliance.

On KITTI-10, our model maintains competitive performance with a minADE of 0.19\,m and a HitRate of 0.96, surpassing MTP and TP in accuracy and consistency. Although our inference time (0.053–0.055\,s) is higher than that of baseline models (0.005–0.018\,s), the additional computation yields 53–62\% improvements in core accuracy metrics, which is a reasonable trade-off in safety-critical applications.

\subsubsection{Qualitative Analysis}

Fig.~\ref{fig:results} presents qualitative examples of predicted trajectories (green) versus ground truth (red). TopoDiffuser demonstrates strong spatial alignment with road geometry across various urban scenes. The model effectively captures multimodal behaviors. Minor deviations occur primarily in highly interactive scenarios, reflecting inherent uncertainty in robot behavior.

\begin{figure}[t!]
    \centering
    \includegraphics[width=0.47\textwidth]{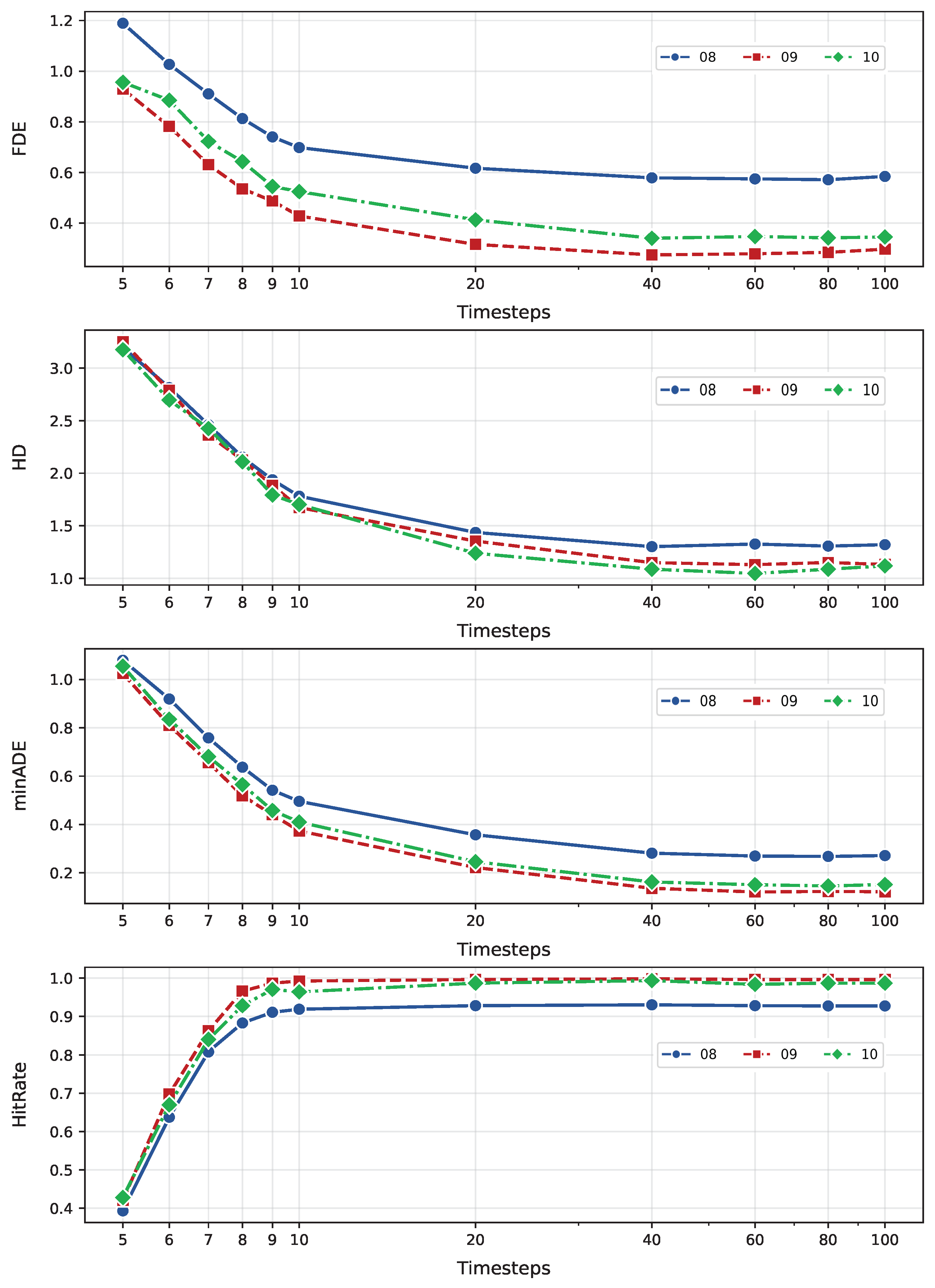}
    \caption{Effect of denoising step count on prediction metrics.}
    \label{fig:step number}
\end{figure}

\begin{figure}[t!]
    \centering
    \includegraphics[width=0.47\textwidth]{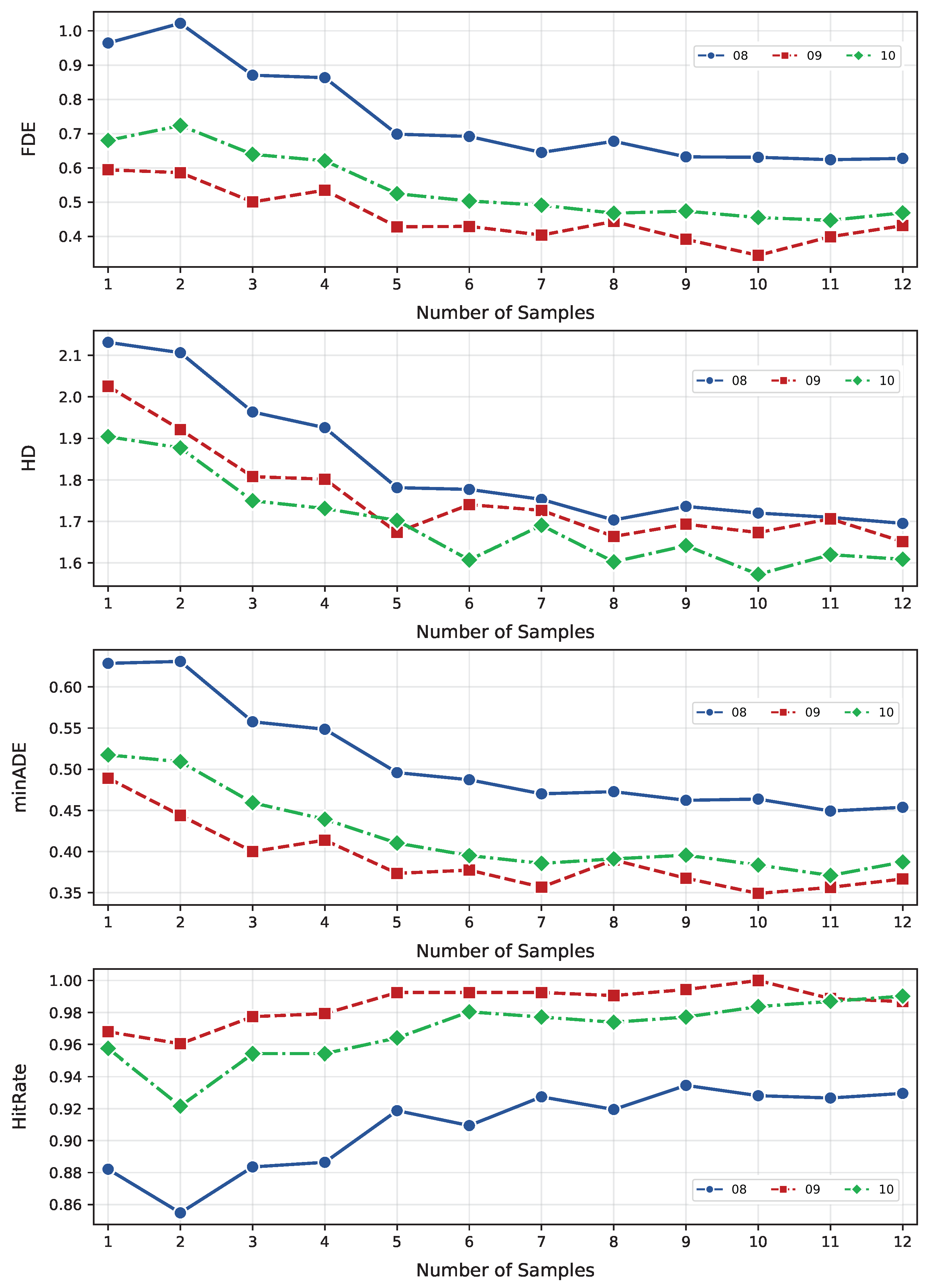}
    \caption{Effect of number of sampled trajectories on prediction accuracy.}
    \label{fig:sample number}
\end{figure}

\begin{table}[t!]
\caption{Ablation results on KITTI-10 showing the impact of different input modalities on prediction performance.}
\small
\centering
\begin{tabular}{lcccc}
    \toprule
    \textbf{Model} & \textbf{FDE} $\downarrow$ & \textbf{minADE} $\downarrow$ & \textbf{HitRate} $\uparrow$ & \textbf{HD} $\downarrow$ \\
    \midrule
    \textbf{L.}                  & 0.55 & 0.28 & 0.93 & 1.55 \\
    \textbf{L. + M.}            & 0.55 & 0.25 & 0.93 & 1.50 \\
    \textbf{L. + M. + H.}       & 0.55 & 0.26 & 0.93 & 1.32 \\
    \bottomrule
\end{tabular}
\label{tab:ablation}
\end{table}

\subsubsection{Ablation Studies}
To investigate the influence of different input modalities on trajectory prediction performance, we conduct ablation studies on the KITTI-10 sequence. The results are summarized in Table~\ref{tab:ablation}.

We begin with a model that uses only the LiDAR (L.) input. This configuration yields an FDE of 0.55\,m and a minADE of 0.28\,m, with a HitRate of 0.93 and HD of 1.55. When incorporating the topometric map (M.), the minADE improves to 0.25\,m and HD decreases to 1.50, confirming that even coarse map priors contribute structural guidance.

Adding the historical trajectory (H.) further refines the prediction. The HD is significantly reduced to 1.32\,m, representing a 14.8\% improvement over the baseline. This indicates that temporal context helps the model better align predictions with the underlying road structure.

% \begin{figure*}[ht]
%     \centering
%     \includegraphics[width=0.95\textwidth]{fig/000019.jpg}
%     \caption{Trajectory refinement over denoising steps.}
%     \label{fig:qualitative}
% \end{figure*}

\subsubsection{Effect of Denoising Steps}
We analyze the impact of the number of denoising steps on prediction performance. As shown in Fig.~\ref{fig:step number}, increasing the number of steps from 5 to 20 improves all metrics. The predicted trajectories become more accurate and better aligned with road constraints.
However, beyond 20 steps, the performance gains tend to saturate, and further increasing the number of steps results in only marginal improvements. This indicates that most of the refinement occurs during the early stages of the denoising process, and using a moderate number of steps is sufficient for practical deployment.

\subsubsection{Effect of Sampling}
We further analyze how the number of sampled trajectories during inference affects prediction performance. As shown in Fig.~\ref{fig:sample number}, increasing the number of samples generally leads to performance improvements across all evaluation metrics.

In particular, when the number of samples increases from 1 to 8, there is a clear downward trend in FDE, minADE, and HD. Meanwhile, the HitRate shows a consistent rise. These trends indicate that drawing more samples allows the model to better capture multimodal uncertainties and generate more accurate and road-compliant trajectories.
However, after reaching around 8 samples, the performance gains begin to saturate. Further increasing the number of samples yields only marginal improvements, suggesting diminishing returns as the sample count grows.

% \subsubsection{Denoising Process Visualization}
% We visualize intermediate trajectory predictions at different denoising steps to illustrate how the diffusion process refines noisy samples into feasible trajectories, as shown in Fig.~\ref{fig:qualitative}.

% In the early steps, trajectories show large deviations and poor alignment with the road. As denoising progresses, they become smoother and increasingly follow the topometric route. By the final step, the predictions align closely with the road geometry and ground-truth paths.

\section{CONCLUSIONS}
In this work, we presented \textit{TopoDiffuser}, a diffusion-based trajectory prediction framework that incorporates topometric maps to address the challenges of multimodality and feasibility in future motion forecasting. By embedding structural guidance into the denoising process, the model generates diverse, realistic, and road-compliant trajectories without relying on explicit constraints. We validated TopoDiffuser through extensive experiments, where it consistently outperformed existing methods. The explicit use of topological information helps ensure that the predicted trajectories align well with road geometry while maintaining the flexibility to represent uncertain driving behaviors. For future work, we plan to enhance the model by integrating perception of dynamic obstacles such as vehicles and pedestrians.

\bibliographystyle{IEEEtran}  
\bibliography{ref} 

\begin{thebibliography}{10}
\providecommand{\url}[1]{#1}
\csname url@samestyle\endcsname
\providecommand{\newblock}{\relax}
\providecommand{\bibinfo}[2]{#2}
\providecommand{\BIBentrySTDinterwordspacing}{\spaceskip=0pt\relax}
\providecommand{\BIBentryALTinterwordstretchfactor}{4}
\providecommand{\BIBentryALTinterwordspacing}{\spaceskip=\fontdimen2\font plus
\BIBentryALTinterwordstretchfactor\fontdimen3\font minus \fontdimen4\font\relax}
\providecommand{\BIBforeignlanguage}[2]{{%
\expandafter\ifx\csname l@#1\endcsname\relax
\typeout{** WARNING: IEEEtran.bst: No hyphenation pattern has been}%
\typeout{** loaded for the language `#1'. Using the pattern for}%
\typeout{** the default language instead.}%
\else
\language=\csname l@#1\endcsname
\fi
#2}}
\providecommand{\BIBdecl}{\relax}
\BIBdecl

\bibitem{Chang2019}
M.-F. Chang, J.~Lambert, P.~Sangkloy, J.~Singh, S.~Bak, A.~Hartnett, D.~Wang, P.~Carr, S.~Lucey, D.~Ramanan, and J.~Hays, ``Argoverse: 3d tracking and forecasting with rich maps,'' in \emph{Proc. IEEE/CVF Conf. Comput. Vis. Pattern Recognit. (CVPR)}, June 2019.

\bibitem{Deo2018}
N.~Deo and M.~M. Trivedi, ``Multi-modal trajectory prediction of surrounding vehicles with maneuver-based lstms,'' in \emph{Proc. IEEE Intell. Vehicles Symp. (IV)}, 2018, pp. 1179--1184.

\bibitem{Ding2019}
W.~Ding, J.~Chen, and S.~Shen, ``Predicting vehicle behaviors over an extended horizon using behavior interaction network,'' in \emph{Proc. IEEE Int. Conf. Robot. Autom. (ICRA)}, 2019, pp. 8634--8640.

\bibitem{Katariya2022}
V.~Katariya, M.~Baharani, N.~Morris, O.~Shoghli, and H.~Tabkhi, ``Deeptrack: Lightweight deep learning for vehicle trajectory prediction in highways,'' \emph{IEEE Trans. Intell. Transp. Syst.}, vol.~23, no.~10, pp. 18\,927--18\,936, 2022.

\bibitem{Li2022}
D.~Li, H.~Li, Y.~Xiao, B.~Li, and B.~Tang, ``Vehicle trajectory prediction for automated driving based on temporal convolution networks,'' in \emph{Proc. WRC Symp. Adv. Robot. Autom. (WRC SARA)}, 2022, pp. 257--262.

\bibitem{Zhang2021}
Y.~Zhang, Y.~Zou, J.~Tang, and J.~Liang, ``Long-term prediction for high-resolution lane-changing data using temporal convolution network,'' \emph{Transportmetrica B: Transport Dyn.}, vol.~10, no.~1, pp. 849--863, Jul. 2021.

\bibitem{Chen2021}
W.~Chen, F.~Wang, and H.~Sun, ``S2tnet: Spatio-temporal transformer networks for trajectory prediction in autonomous driving,'' in \emph{Proc. 13th Asian Conf. Mach. Learn. (ACML)}, vol. 157.\hskip 1em plus 0.5em minus 0.4em\relax PMLR, Nov. 17--19 2021, pp. 454--469.

\bibitem{Liu2021}
Y.~Liu, J.~Zhang, L.~Fang, Q.~Jiang, and B.~Zhou, ``Multimodal motion prediction with stacked transformers,'' in \emph{Proc. IEEE/CVF Conf. Comput. Vis. Pattern Recognit. (CVPR)}, June 2021, pp. 7577--7586.

\bibitem{Quintanar2021}
A.~Quintanar, D.~Fernández-Llorca, I.~Parra, R.~Izquierdo, and M.~A. Sotelo, ``Predicting vehicles trajectories in urban scenarios with transformer networks and augmented information,'' in \emph{Proc. IEEE Intell. Vehicles Symp. (IV)}, 2021, pp. 1051--1056.

\bibitem{Chandra2020}
R.~Chandra, T.~Guan, S.~Panuganti, T.~Mittal, U.~Bhattacharya, A.~Bera, and D.~Manocha, ``Forecasting trajectory and behavior of road-agents using spectral clustering in graph-lstms,'' \emph{IEEE Robot. Autom. Lett.}, vol.~5, no.~3, pp. 4882--4890, 2020.

\bibitem{Li2019}
X.~Li, X.~Ying, and M.~C. Chuah, ``Grip: Graph-based interaction-aware trajectory prediction,'' in \emph{Proc. IEEE Intell. Transp. Syst. Conf. (ITSC)}, 2019, pp. 3960--3966.

\bibitem{Fu2022}
M.~Fu, T.~Zhang, W.~Song, Y.~Yang, and M.~Wang, ``Trajectory prediction-based local spatio-temporal navigation map for autonomous driving in dynamic highway environments,'' \emph{IEEE Trans. Intell. Transp. Syst.}, vol.~23, no.~7, pp. 6418--6429, 2022.

\bibitem{Meng2021}
Q.~Meng, B.~Shang, Y.~Liu, H.~Guo, and X.~Zhao, ``Intelligent vehicles trajectory prediction with spatial and temporal attention mechanism,'' \emph{IFAC-PapersOnLine}, vol.~54, no.~10, pp. 454--459, 2021.

\bibitem{Guo2023}
H.~Guo, Q.~Meng, X.~Zhao, J.~Liu, D.~Cao, and H.~Chen, ``Map-enhanced generative adversarial trajectory prediction method for automated vehicles,'' \emph{Inf. Sci.}, vol. 622, pp. 1033--1049, 2023.

\bibitem{Hegde2020}
C.~Hegde, S.~Dash, and P.~Agarwal, ``Vehicle trajectory prediction using gan,'' in \emph{Proc. Int. Conf. I-SMAC (IoT Soc. Mobile Analytics Cloud)}, 2020, pp. 502--507.

\bibitem{Li2021}
X.~Li, G.~Rosman, I.~Gilitschenski, C.-I. Vasile, J.~A. DeCastro, S.~Karaman, and D.~Rus, ``Vehicle trajectory prediction using generative adversarial network with temporal logic syntax tree features,'' \emph{IEEE Robot. Autom. Lett.}, vol.~6, no.~2, pp. 3459--3466, 2021.

\bibitem{Bhattacharyya2020}
\BIBentryALTinterwordspacing
A.~Bhattacharyya, M.~Hanselmann, M.~Fritz, B.~Schiele, and C.-N. Straehle, ``Conditional flow variational autoencoders for structured sequence prediction,'' \emph{arXiv preprint arXiv:1908.09008}, 2020. [Online]. Available: \url{https://arxiv.org/abs/1908.09008}
\BIBentrySTDinterwordspacing

\bibitem{Cho2019}
K.~Cho, T.~Ha, G.~Lee, and S.~Oh, ``Deep predictive autonomous driving using multi-agent joint trajectory prediction and traffic rules,'' in \emph{Proc. IEEE/RSJ Int. Conf. Intell. Robots Syst. (IROS)}, 2019, pp. 2076--2081.

\bibitem{Dulian2021}
\BIBentryALTinterwordspacing
A.~Dulian and J.~C. Murray, ``Multi-modal anticipation of stochastic trajectories in a dynamic environment with conditional variational autoencoders,'' \emph{arXiv preprint arXiv:2103.03912}, 2021. [Online]. Available: \url{https://arxiv.org/abs/2103.03912}
\BIBentrySTDinterwordspacing

\bibitem{Bahram2016}
M.~Bahram, A.~Lawitzky, J.~Friedrichs, M.~Aeberhard, and D.~Wollherr, ``A game-theoretic approach to replanning-aware interactive scene prediction and planning,'' \emph{IEEE Trans. Veh. Technol.}, vol.~65, no.~6, pp. 3981--3992, 2016.

\bibitem{Gao2020}
J.~Gao, C.~Sun, H.~Zhao, Y.~Shen, D.~Anguelov, C.~Li, and C.~Schmid, ``Vectornet: Encoding hd maps and agent dynamics from vectorized representation,'' in \emph{Proc. IEEE/CVF Conf. Comput. Vis. Pattern Recognit. (CVPR)}, 2020.

\bibitem{Gu2021}
J.~Gu, C.~Sun, and H.~Zhao, ``Densetnt: End-to-end trajectory prediction from dense goal sets,'' in \emph{Proc. IEEE/CVF Int. Conf. Comput. Vis. (ICCV)}, 2021, pp. 15\,303--15\,312.

\bibitem{Liang2020}
M.~Liang, B.~Yang, R.~Hu, Y.~Chen, R.~Liao, S.~Feng, and R.~Urtasun, ``Learning lane graph representations for motion forecasting,'' in \emph{Proc. Eur. Conf. Comput. Vis. (ECCV)}.\hskip 1em plus 0.5em minus 0.4em\relax Berlin, Heidelberg: Springer, 2020, pp. 541--556.

\bibitem{Zeng2021}
W.~Zeng, M.~Liang, R.~Liao, and R.~Urtasun, ``Lanercnn: Distributed representations for graph-centric motion forecasting,'' in \emph{Proc. IEEE/RSJ Int. Conf. Intell. Robots Syst. (IROS)}, 2021, pp. 532--539.

\bibitem{Wang2023}
Z.~Wang, J.~Guo, Z.~Hu, H.~Zhang, J.~Zhang, and J.~Pu, ``Lane transformer: A high-efficiency trajectory prediction model,'' \emph{IEEE Open J. Intell. Transp. Syst.}, vol.~4, pp. 2--13, 2023.

\bibitem{Zeng2019}
W.~Zeng, W.~Luo, S.~Suo, A.~Sadat, B.~Yang, S.~Casas, and R.~Urtasun, ``End-to-end interpretable neural motion planner,'' in \emph{Proc. IEEE/CVF Conf. Comput. Vis. Pattern Recognit. (CVPR)}.\hskip 1em plus 0.5em minus 0.4em\relax IEEE, Jun. 2019, pp. 8660--8669.

\bibitem{Xu2022}
J.~Xu, L.~Xiao, D.~Zhao, Y.~Nie, and B.~Dai, ``Trajectory prediction for autonomous driving with topometric map,'' in \emph{Proc. IEEE Int. Conf. Robot. Autom. (ICRA)}.\hskip 1em plus 0.5em minus 0.4em\relax IEEE, May 2022, pp. 8403--8408.

\bibitem{Geiger:2012}
A.~Geiger, P.~Lenz, and R.~Urtasun, ``Are we ready for autonomous driving? the kitti vision benchmark suite,'' in \emph{CVPR}, Washington, DC, USA, 2012.

\bibitem{2019CoverNet}
T.~Phan-Minh, E.~C. Grigore, F.~A. Boulton, O.~Beijbom, and E.~M. Wolff, ``Covernet: Multimodal behavior prediction using trajectory sets,'' in \emph{CVPR}, 2019.

\bibitem{2018Multimodal}
H.~Cui, V.~Radosavljevic, F.~C. Chou, T.~H. Lin, T.~Nguyen, T.~K. Huang, J.~Schneider, and N.~Djuric, ``Multimodal trajectory predictions for autonomous driving using deep convolutional networks,'' in \emph{ICRA}, 2018.

\end{thebibliography}
\end{document}